\documentclass[letterpaper, 10pt, journal, twoside]{IEEEtran}

\usepackage{cite}
\usepackage{amsmath,amssymb,amsfonts}
\usepackage{algorithmic}
\usepackage{graphicx,float}
\usepackage{textcomp}
\usepackage{xcolor}

\makeatletter
\newcommand{\linebreakand}{%
  \end{@IEEEauthorhalign}
  \hfill\mbox{}\par
  \mbox{}\hfill\begin{@IEEEauthorhalign}
}
\makeatother

\usepackage[acronym]{glossaries}
\newacronym{ros}{ROS}{Robot Operating System}
\newacronym{emaf}{EMAF}{Exponential Moving Average Filter}
\newacronym{csv}{CSV}{Comma-Separated Values}
\newacronym{mhsa}{MHSA}{Multi-Head Self-Attention}
\newacronym{ffn}{FFNN}{Feed-Forward Neural Network}
\newacronym{tf}{TF}{Force-Torque Sensor}
\newacronym{ur3e}{UR3e}{Universal Robots UR3e}
\newacronym{acc}{ACC}{Accuracy}
\newacronym{prec}{Prec}{Precision}
\newacronym{cps}{CPS}{Cyber-Physical System}
\newacronym{dtw}{DTW}{Dynamic Time Warping}
\newacronym{ai}{AI}{Artificial Intelligence}
\newacronym{gdpr}{GDPR}{General Data Protection Regulation}

\makeatletter
\long\def\@makecaption#1#2{\ifx\@captype\@IEEEtablestring%
\footnotesize\begin{center}{\normalfont\footnotesize #1}\\
{\normalfont\footnotesize\scshape #2}\end{center}%
\@IEEEtablecaptionsepspace
\else
\@IEEEfigurecaptionsepspace
\setbox\@tempboxa\hbox{\normalfont\footnotesize {#1.}~~ #2}%
\ifdim \wd\@tempboxa >\hsize%
\setbox\@tempboxa\hbox{\normalfont\footnotesize {#1.}~~ }%
\parbox[t]{\hsize}{\normalfont\footnotesize \noindent\unhbox\@tempboxa#2}%
\else
\hbox to\hsize{\normalfont\footnotesize\hfil\box\@tempboxa\hfil}\fi\fi}
\makeatother

\begin{document}
\title{Haptic-Based User Authentication for Tele-robotic
System}

% \author{Michael~Shell,~\IEEEmembership{Member,~IEEE,}
%         John~Doe,~\IEEEmembership{Fellow,~OSA,}
%         and~Jane~Doe,~\IEEEmembership{Life~Fellow,~IEEE}% <-this % stops a space
% \thanks{M. Shell was with the Department
% of Electrical and Computer Engineering, Georgia Institute of Technology, Atlanta,
% GA, 30332 USA e-mail: (see http://www.michaelshell.org/contact.html).}% <-this % stops a space
% \thanks{J. Doe and J. Doe are with Anonymous University.}% <-this % stops a space
% \thanks{Manuscript received April 19, 2005; revised August 26, 2015.}}
\author{Rongyu Yu$^{1}$, Kan Chen$^{2}$, Zeyu Deng$^{3}$, Chen Wang$^{3}$, Burak Kizilkaya$^{2}$, Emma Li$^{2}$
\thanks{This work involved human subjects in its research. Approval of all ethical and experimental procedures and protocols was granted by the University Ethics Committee of the University of Glasgow under Application 300230225 and performed in line with European General Data Protection Regulation (GDPR).}
\thanks{$^{1}$Rongyu Yu is with James Watt School of Engineering, University of Glasgow, Glasgow, G12 8QQ, UK
        {\tt\footnotesize 2658366y@student.gla.ac.uk, burak.kizilkaya@glasgow.ac.uk}}%
\thanks{$^{2}$Kan Chen, Burak Kizilkaya, Emma Li are with the School of Computing Science, University of Glasgow, Glasgow, G12 8RZ, UK
        {\tt\footnotesize liying.li@glasgow.ac.uk, k.chen.1@research.gla.ac.uk}}%
\thanks{$^{3}$Zeyu Deng and Chen Wang are with the Computer Science Department, Southern Methodist University, Dallas, TX, United States
        {\tt\footnotesize zeyud@smu.edu, cwang6@smu.edu}}%
        }

\maketitle


\begin{abstract}

Tele-operated robots rely on real-time user behavior mapping for remote tasks, but ensuring secure authentication remains a challenge. Traditional methods, such as passwords and static biometrics, are vulnerable to spoofing and replay attacks, particularly in high-stakes, continuous interactions. This paper presents a novel anti-spoofing and anti-replay authentication approach that leverages distinctive user behavioral features extracted from haptic feedback during human–robot interactions. To evaluate our authentication approach, we collected a time-series force feedback dataset from 15 participants performing seven distinct tasks. We then developed a transformer-based deep learning model to extract temporal features from the haptic signals. By analyzing user-specific force dynamics, our method achieves over 90\% accuracy in both user identification and task classification, demonstrating its potential for enhancing access control and identity assurance in tele-robotic systems.

\end{abstract}

% \begin{IEEEkeywords}
% IEEE, IEEEtran, journal, \LaTeX, paper, template.
% \end{IEEEkeywords}

% \fi
\IEEEpeerreviewmaketitle

\section{Introduction}

%Despite recent advances in cognitive robotics, 

Cognitive robotics has made significant progress in recent years, such as deep learning-based perception for real-time object recognition and grasping \cite{yin2021modeling}, as well as reinforcement learning techniques \cite{kober2013reinforcement} enabling robots to autonomously acquire complex motor skills. However, current fully autonomous systems still struggle to replicate nuanced, human-like decision-making and dexterous manipulation, that are particularly essential for many mission-critical tasks such as remote surgery \cite{ballantyne2002robotic} and hazardous material handling \cite{tokatli2021robot}, where expert human judgment and direct control are crucial for ensuring safety and security \cite{darvish2023teleoperation}.

At the same time, as robots become more integrated into our daily lives, robust remote access control and user authentication are essential to prevent unauthorized access, secure sensitive data, and guard against potential cyber threats. Weak security measures pose a significant risk, allowing adversary attackers to manipulate operations, steal information, or disrupt services, thereby highlighting the need for stringent safeguards and secure interfaces in both networked robotic systems and human–robot interactions.

Haptic feedback play a crucial role in improving user interaction with robotic systems by providing tactile responses that enhance control precision, enabling operators to perceive physical phenomena such as force, vibration, and impact in both virtual and remote environments. This real-time tactile perception not only heightens immersion and realism, but also provides critical cues for precise manipulation, enhanced safety, and efficient task execution.

\gls{cps} \cite{baheti2011cyber} merge computing, networking, and physical components to enable real-time monitoring and control across both digital and physical domains. By seamlessly linking devices, sensors, and human operators, \gls{cps} incorporate advanced feedback loops that significantly enhance processes such as human–robot interactions, providing lower-latency and higher-fidelity data \cite{schirner2013future}. In the field of robotics, digital twins leverage these feedback loops to enable more precise coordination between operators and robots, support complex decision-making, and facilitate tasks such as remote surgery or autonomous navigation. However, this high degree of connectivity also increases security risks \cite{khalid2018security}, as malicious actors may exploit vulnerable communication channels or compromised devices to manipulate system behavior. Consequently, \gls{cps} require robust user authentication and access-control mechanisms to block unauthorized access, protect data integrity, and ensure safe, reliable operation.

Traditional credential-based authentication methods for robotic security are often vulnerable to phishing, brute-force attacks, and shoulder surfing. For example, passwords can be easily stolen, compromised, or forgotten \cite{bonneau2012quest}. By contrast, biometric techniques offer stronger protection against spoofing by leveraging unique human physical traits, which are inherently more difficult to replicate or forge.

These approaches can be broadly categorized into physiological and behavioral biometrics. Physiological biometrics utilize personal physical attributes such as fingerprints, facial features, and iris patterns. Behavioral biometrics, on the other hand, analyze distinctive human behavioral patterns such as typing rhythms, signature dynamics, and gait. Behavioral biometrics are challenging to replicate and offer the additional benefit of enabling continuous authentication \cite{jain2006biometrics}.

Haptic-based behavioral biometrics provide a dynamic, hard-to-forge, and context-sensitive authentication technique by utilizing each individual's distinct force dynamics \cite{el2007novel}. Generally, this approach is applied as the additional layer of security (two-factor authentication) to traditional credential-based authentication techniques, such as PINs \cite{bhole2024haptic2fa,bianchi2010secure}, signatures \cite{dhandapani2021hapticlock}, or pattern locks \cite{yan2015haptic}, which are often compromised through “shoulder surfing,” where attackers observe passwords without the user’s consent.

In \cite{bhole2024haptic2fa}, authors introduce an eyes-free mobile authentication method using random starting digits, vertical swipe gestures, and morse code vibration feedback, validated via a 20-participant user study and a 15-participant shoulder-surfing experiment. In \cite{bianchi2010secure}, Bianchi et al.\ propose the Secure Haptic Keypad (SHK), a tactile-based PIN entry system that mitigates shoulder-surfing attacks by encoding each digit as a unique vibration pattern. User studies show that SHK provides enhanced security with minimal impact on input speed and accuracy. In \cite{dhandapani2021hapticlock}, the authors leverage a 6-DOF Phantom Omni Device to record multi-dimensional input data (position, velocity, forces, pen orientation) and uses a \gls{dtw} to extract features for an artificial neural network, achieving high accuracy, resisting forgery, and remaining user-friendly.  Furthermore, in \cite{yan2015haptic}, the authors presents a dial-based interface for public terminals (e.g., ATMs) that employs tactons (structured vibration patterns) to improve PIN entry efficiency and reduce errors.

Furthermore, earlier work \cite{huang2021toward} has demonstrated that each user exhibits a distinctive and identifiable motion pattern when teleoperating a robotic system, pointing to a potential behavioral-biometric solution for robotic security. Meanwhile, robot learning from demonstration (LfD) is a well-established method that allows robots to learn to replicate human behaviors by observing demonstrations. Our previous research also explored how robots can learn user-specific keystroke dynamics \cite{keystroke}. Building on these insights, it is promising to incorporate user-specific force dynamics in future studies, enabling robots to capture the specialized behaviors of different task experts.

In summary, existing haptic biometric methods reinforce traditional authentication by leveraging each user’s unique force dynamics, offering an additional layer of security. This approach provides robust defense against shoulder-surfing attacks and shows strong potential to improve usability, accuracy, and resilience against impersonation threats. However, as tele-operated robotic arms become more common in daily life, the need for robust, continuous authentication during teleoperation grows. Current solutions rarely apply behavioral biometrics throughout the robotic manipulation process. To address this gap, we propose a novel, haptic-based user authentication system for tele-robotic applications, which uses user-specific force feedback patterns to provide continuous, reliable verification. To our knowledge, this is the first work demonstrating the feasibility of using haptic biometrics to authenticate tele-operators.

The main contributions can be summarized as follows.

\begin{enumerate}
    \item In this paper, we develop a haptic-based user authentication system for tele-robotic applications. We show that force feedback inherently carries personal information about the operator in human–machine interaction, especially under remote control.
    \item We conducted a large-scale user data collection, gathering 120 samples across 7 tasks from 15 participants. Our findings demonstrate that haptic signals during robot teleoperation can be leveraged for user identification and task classification.%, although they also raise important security and privacy concerns.
    \item We propose a transformer-based deep learning algorithm that extracts time-series features from haptic signals for user identification and task classification, using the data set we collected.
    \item Finally, we validate our approach by collecting data from 15 participants using a virtual motion-controlled robotic arm platform. Each participant performs seven tasks by writing the letters \emph{a, b, c, d, e, f, g} on a plate. We then evaluate both the raw and filtered force data, demonstrating that haptic signals can achieve over 90\% accuracy in user identification and task classification.
\end{enumerate}

The remainder of this paper is structured as follows. 
In Section~\ref{sec:System Architecture and Experimental Procedure}, 
we describe the system architecture and experimental procedure. 
Section~\ref{sec:User Authentication Design} introduces the design 
and implementation of our user authentication mechanism, including 
the data processing of haptic signals and details of model training. 
Section~\ref{sec:Performance Evaluation} presents the results and 
provides a thorough analysis. Finally, in Section~\ref{sec:Conclusion}, 
we conclude the study by summarizing the key findings and discussing their implications for future research.

\section{System Architecture and Experimental Procedure}
\label{sec:System Architecture and Experimental Procedure}

\subsection{System Overview}

\begin{figure}[!t]
  \centering
  % 调整下面 trim={} 中的数值以满足裁边需求。单位可为 cm 或 mm
  \includegraphics[width=0.85\linewidth,trim={1cm 1cm 1cm 1cm},clip]{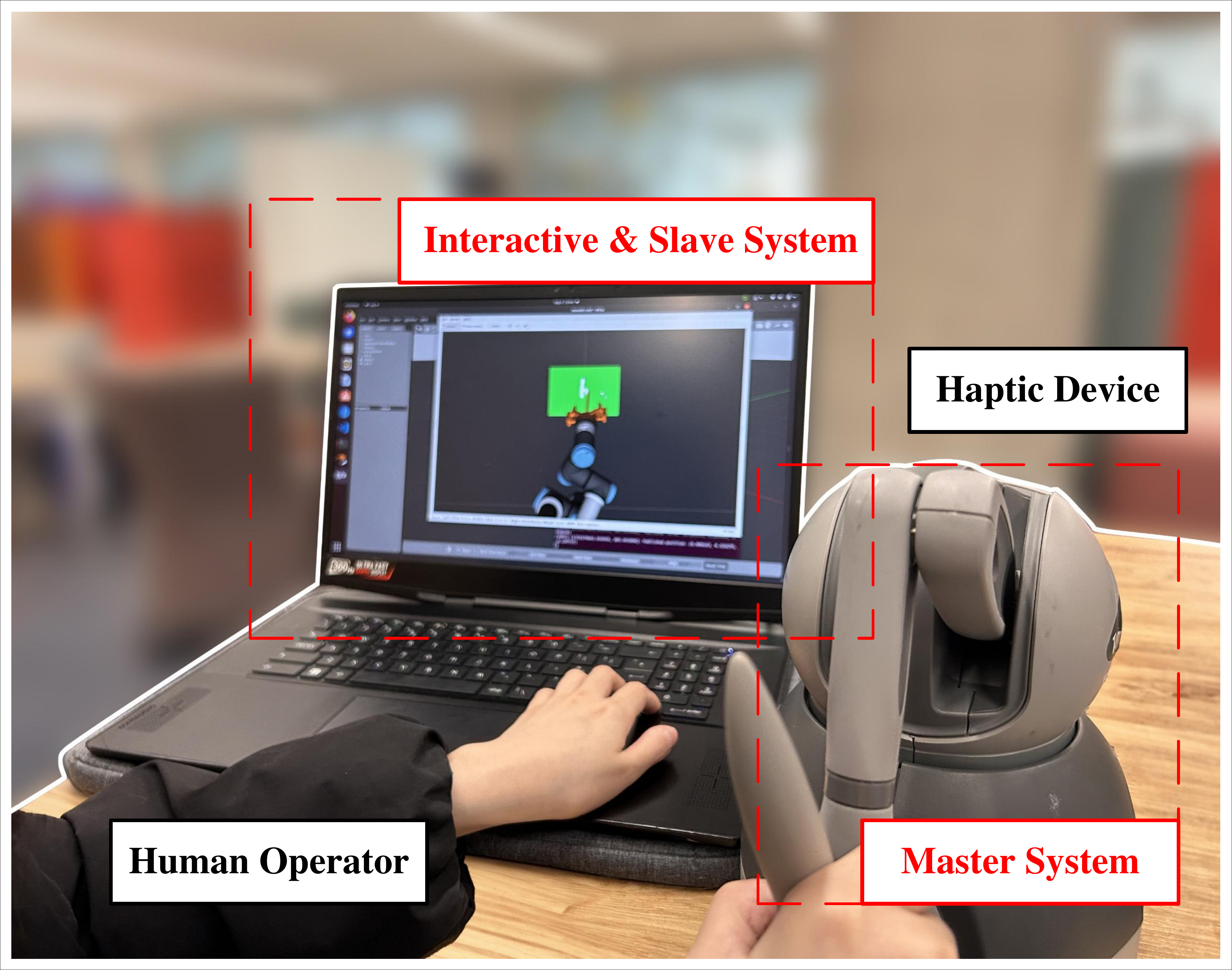}
  \caption{Overview of the virtual teleoperation system illustrating the four main components. A human operator (left) manipulates the master systems via the haptic device (right), which provides haptic force feedback. These inputs are transmitted to the interactive and slave systems displayed on the laptop, enabling intuitive remote control and interaction.}
  \label{fig6:experimental setup}
\end{figure}

The experimental system in this study comprises both hardware and software components that together enable precise simulation and data recording of human-robot interactions. As illustrated in Fig. \ref{fig6:experimental setup}, a human operator interacts with a virtual robot using a haptic device controller, which provides real-time force feedback. These interactions are transmitted over an virtual tele-robotic system, facilitating intuitive remote control and operation. 

%\noindent \textbf{\textcolor{black}{Software Component.}}
\subsection{Software Components}
\label{subsec:software_components}

In our system, the robotic virtual environment and control infrastructure are built using 
the Robot Operating System (ROS) Noetic \texttt{MoveIt} framework \cite{wikiROS}. 
This platform integrates motion planning, kinematics solving, collision detection, and visualization 
into a unified architecture. We employ Gazebo \cite{classic_gazebo} rendering engine to construct the virtual environment, as it offers relatively low CPU latency and delivers accurate sensing capabilities along with real-time haptic force feedback. We will use RViz\cite{rviz_wiki}  for both object visualization and handwritten letter visualization. 

\subsubsection{Virtual Environment and Update Rates}
The Gazebo simulation environment is configured with a $250$\,Hz update rate for both the virtual world and the robot state controller, ensuring responsive control and seamless interaction.

All data streams including raw data from \gls{tf} and filtered force data are recorded at 250\,Hz to ensure consistency. By aligning each component (simulation, controller, and data recording) to the same update frequency, we achieve stable control cycles and synchronized sensor data acquisition.

\subsection{Haptic Rendering}  

We employ a simple haptic rendering mechanism that reads and scales raw force data from \gls{tf}. When the pen tip contacts the plate, \gls{tf} collision detection is triggered. An \gls{emaf} with a smoothing constant \(\alpha = 0.001\) is then applied to update the force, and the smoothed force is provided to the user as haptic feedback, ensuring a seamless and responsive virtual interaction. If the detected raw force falls below a preset threshold (for example, when the user's pen tip leaves the platform), the force quickly decays.

%, making t setup well-suited for advanced robotic control and real-time evaluation.

%\noindent \textbf{\textcolor{black}{Hardware Component.}} 
\subsubsection{Hardware Components}

 We employ the \textit{Touch Haptic} device (also known as the \textit{Geomagic Touch}) as the human input interface. This controller provides six degrees of freedom (DoF) for tracking the spatial position and orientation of the user’s hand, enabling intuitive manipulation of virtual robotic arms in the world space. It also delivers three-dimensional force feedback for interactive and natural control. Additionally, we use an MSI GS77 laptop configured with a 12th Gen Intel i9-12900H processor to simulate the virtual environment and robot, as well as to run an interactive master–slave system for real-time robot manipulation via the haptic device.

\subsection{Experimental Procedure and Data Collection}\label{experiment}

\begin{figure}[!t]
  \centering
  \vspace{2mm}
  \includegraphics[width=0.95\linewidth]{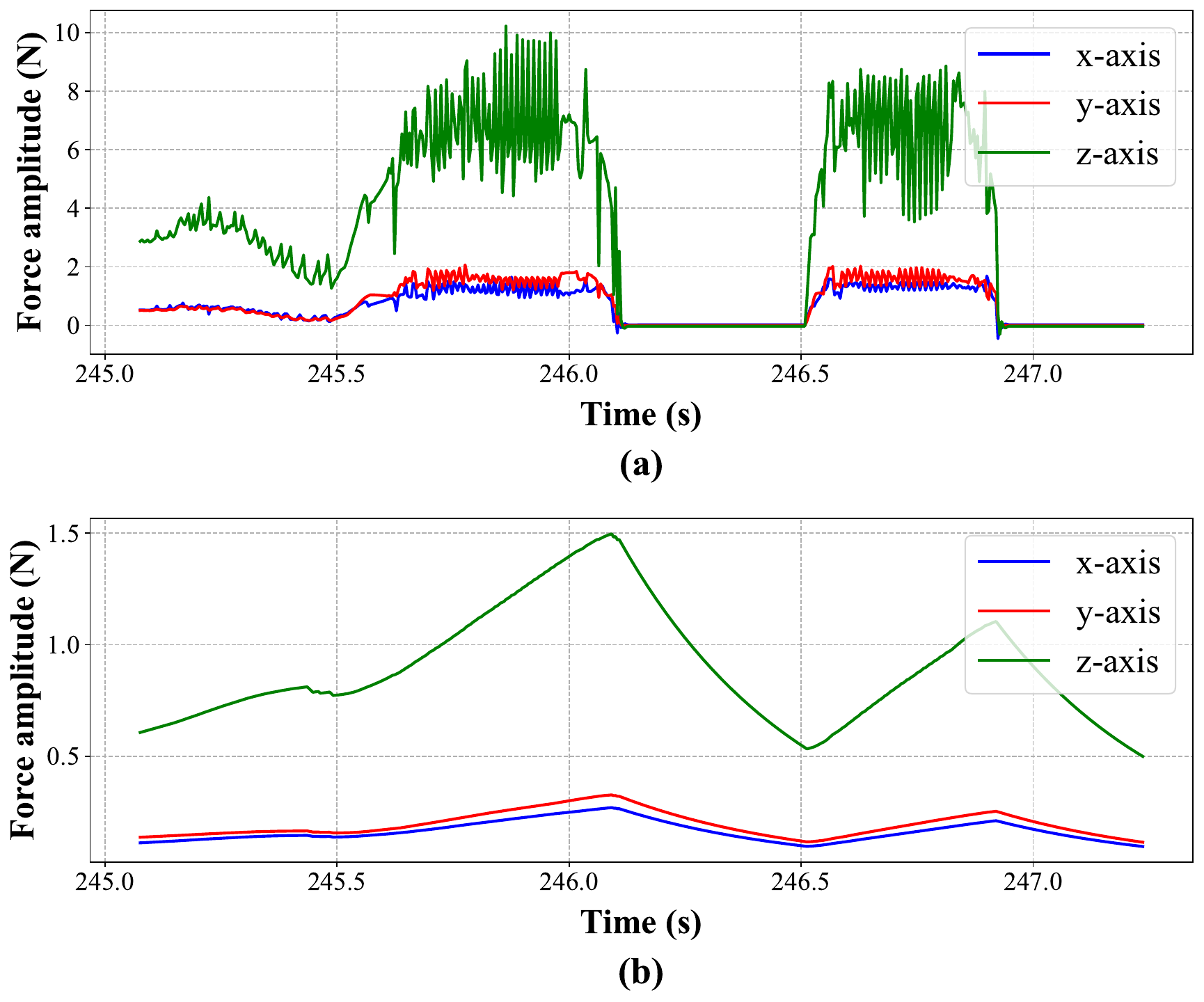}
  \caption{%
    Illustrations of the force signals (a) before filtering and (b) after applying a \gls{emaf} across the x-axis, y-axis, and z-axis. The raw signals in (a) capture the full 
    range of force fluctuations, while (b) shows the smoothed force profiles highlighting 
    the major variations over time.
  }
  \label{fig:combined_force}
\end{figure}

The hand movements of the participants were directly assigned to the end effector of the robotic arm. Data collected from 15 volunteers produced a total of \(120 \times 7 \times 15 = 12{,}600\) samples. Data collection was organized into seven sessions. A short break followed each session, and participants could request additional pauses at any time. All trials were carried out in the school library under natural conditions of ambient.

The force data in the \(x\)-, \(y\)-, and \(z\)-directions, along with the corresponding virtual timestamps, were saved to a \gls{csv} file, anonymized, and stored in compliance with \gls{gdpr} requirements. To preserve realistic interaction scenarios, we deliberately applied no noise filtering or other preprocessing to the haptic data. All subsequent experiments and analyses relied on this unfiltered dataset, ensuring that the results are representative of practical application environments.

As shown in Fig \ref{fig:combined_force}, during the experiment, \textbf{two types of force data} 
were recorded:
\begin{enumerate}
    \item \textbf{Raw Force Data:} The force data were recorded from the force torque sensor of \gls{ur3e}.
    
    \item \textbf{Filtered Force Data:} The data were collected by applying an \gls{emaf} to the raw force data.
\end{enumerate}

%\noindent \textbf{\textcolor{black}{Data Recording.}}
%\subsection{Data Collection}
%When the user begins to write letters on the plate, they are instructed to start at the center for all beginning of letter, which is on a green flat plate marked with a black icon placed on the surface. When the participant presses the laptop’s left arrow key, both data recording and letter visualization are triggered. Pressing the right arrow key stops the data recording and removes the letter from the screen. The \gls{tf} in Wrist 3 of robot detects the external loads generated by this collision in real time. Additionally, the real-time MoveIt Servo module is employed to convert high-level end-effector position commands into low-level joint positions, enabling responsive and precise motion control based on user intention.

All participants signed informed consent forms and all collected data, including personal information, were processed in strict accordance with the university’s ethics regulations. To protect participant privacy, all identifiers were anonymized before storage and data was kept on secure, access-controlled servers. Biometric identifiers or personally traceable metadata were not retained. Participants were explicitly informed of the purpose of data collection, the scope of data usage, and their right to withdraw at any time without penalty. Furthermore, the study design followed the GDPR principles of data minimization and purpose limitation, ensuring that only the data strictly necessary for the research objectives were collected and retained.

\section{User Authentication Design}
\label{sec:User Authentication Design}

\begin{figure*}[!t]
  \centering
  \vspace{2mm}
  % 尝试适当增加裁切边距，以去除灰色描边
  \includegraphics[width=0.9\textwidth,trim={5mm 5mm 5mm 5mm},clip]{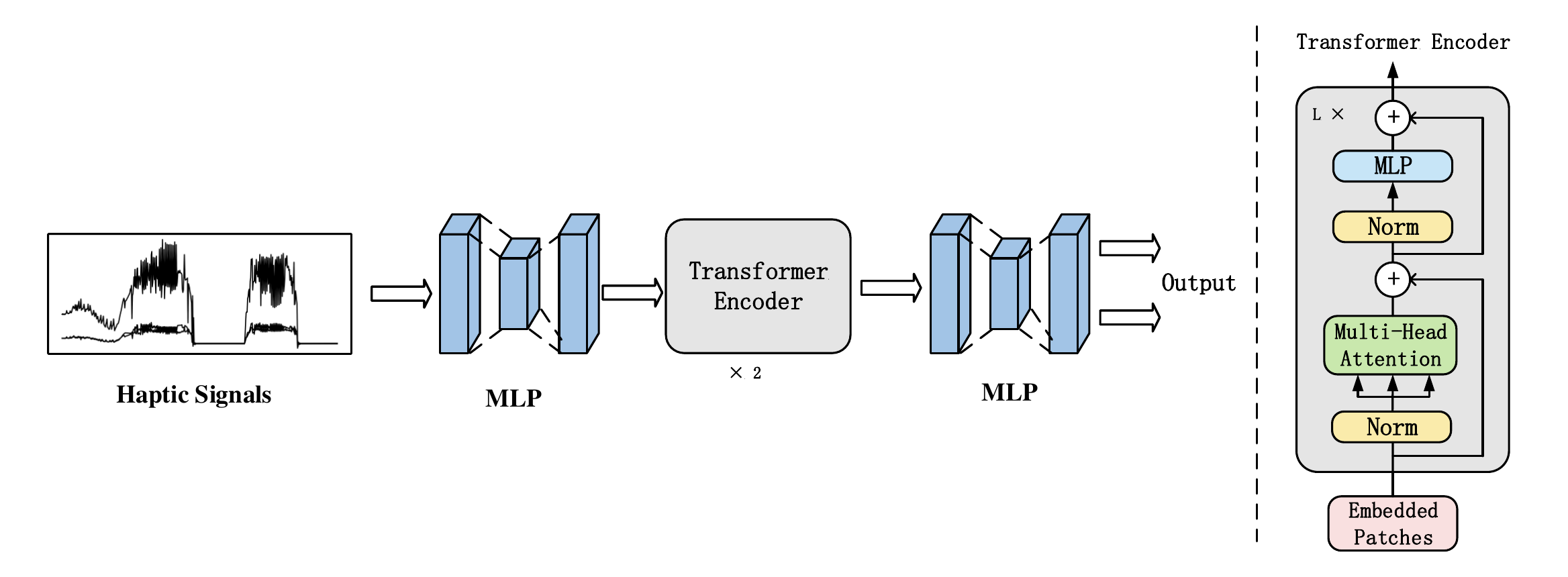}
  \caption{
    An overview of the proposed transformer-based model.}
  \label{fig:model_architecture}
\end{figure*}

\subsection{Data Processing}
\label{sec:Data Processing}

%\noindent \textbf{\textcolor{black}{Feature Extraction.}} 
\subsubsection{Feature Extraction}
We record three-axis force measurements \(\{F_x(t), F_y(t), F_z(t)\}\) at a sampling rate \(S = 250\,\mathrm{Hz}\). For each pair of consecutive time samples, we compute higher-order force features by 
subtracting adjacent sample values and multiplying by \(S\). Specifically, we derive the instantaneous force difference, force velocity, force acceleration, and force jerk. As summarized in Table~\ref{tab:force_features}, each pair of adjacent data points thus yields 13 derived features.

\begin{table}[b]
\centering
\scriptsize
\caption{Force Vector Based Feature Extraction}
\label{tab:force_features}
\begin{tabular}{lp{5.5cm}}
\hline
\textbf{Feature} & \textbf{Notation / Definition} \\
\hline

\textbf{Force Difference} 
& 
\(
\Delta \mathbf{F}(t) 
= \|\mathbf{F}(t+1) - \mathbf{F}(t)\|
\)
\\[0.8em]

\textbf{Velocity} 
& 
\(
\dot{\mathbf{F}}(t) 
= \bigl(\dot{F}_x(t),\, \dot{F}_y(t),\, \dot{F}_z(t)\bigr)^T
\)
\\[0.8em]

\textbf{Velocity Norm} 
& 
\(
\|\dot{\mathbf{F}}(t)\|
= \sqrt{\dot{F}_x^2(t) + \dot{F}_y^2(t) + \dot{F}_z^2(t)}
\)
\\[0.8em]

\textbf{Acceleration} 
& 
\(
\ddot{\mathbf{F}}(t) 
= \bigl(\ddot{F}_x(t),\, \ddot{F}_y(t),\, \ddot{F}_z(t)\bigr)^T
\)
\\[0.8em]

\textbf{Acceleration Norm} 
& 
\(
\|\ddot{\mathbf{F}}(t)\|
= \sqrt{
  \ddot{F}_x^2(t)
+ \ddot{F}_y^2(t)
+ \ddot{F}_z^2(t)
}
\)
\\[0.8em]

\textbf{Jerk} 
& 
\(
\dddot{\mathbf{F}}(t)
= \bigl(\dddot{F}_x(t),\, \dddot{F}_y(t),\, \dddot{F}_z(t)\bigr)^T
\)
\\[0.8em]

\textbf{Jerk Norm}
&
\(
\|\dddot{\mathbf{F}}(t)\|
= \sqrt{
  \dddot{F}_x^2(t)
+ \dddot{F}_y^2(t)
+ \dddot{F}_z^2(t)
}
\)
\\[0.8em]

\hline
\end{tabular}
\end{table}

\noindent

\subsection{Model Training}

%\noindent \textbf{\textcolor{black}{Model Architecture.}}
\subsubsection{Model Architecture}
As shown in Figure~\ref{fig:model_architecture}, we employ a two-layer Transformer-based model to classify haptic signals. The Transformer encoder is composed of multiple stacked encoder layers, each consisting of two main components: \gls{mhsa} and \gls{ffn}. In the multi-head self-attention mechanism, the input haptic signal sequence is first mapped into \textit{Query}, \textit{Key}, and \textit{Value} vectors. By computing attention weights among different time steps or features in the sequence, the relevant information is aggregated, enabling the model to capture critical temporal and feature dependencies in the haptic data. Subsequently, the feed-forward network is applied independently to each position in the sequence and contains two fully connected layers with a ReLU activation in between. The first layer projects the features into a higher-dimensional space, and the second layer maps them back to the original model dimension.

%\noindent \textbf{\textcolor{black}{Training Procedure.}}
\subsubsection{Training Procedure}
In this study, we trained separate models for task classification and user identification.
We begin by randomly shuffling the entire dataset. 
Then, for each user performing a given task, we split their data into 100 samples for training and 20 samples for testing.

For user identification, we trained 7 task-specific models, one per letter, where each user provides 100 training samples (1,500 total) and 20 testing samples (300 total) for that letter. 
For task classification, we trained 15 user-specific models, each with 100 training samples per letter (700 total), and then tested them on 7 letters, with 20 samples per letter (140 total).

%\noindent \textbf{\textcolor{black}{Implementation Details.}}
\subsubsection{Implementation Details}
For user identification and task classification, we adopt slightly different approaches to process
the force data.  For task classification, we downsample the extracted force features to a fixed length of 64. By contrast, for user identification, we downsample each sequence to a length of 512, providing a richer temporal context that helps distinguish subtle individual-specific force signatures. Our model is implemented using PyTorch and PyTorch Lightning for efficient training, with the Adam optimizer at a learning rate of \(10^{-4}\) and a cosine annealing scheduler to mitigate overfitting, all run for 100 epochs for both user identification 
and task classification purposes. Both training tasks use a batch size of 16. Additionally, each Transformer block features a hidden dimension of 256 and 16 attention heads, along with a feed-forward sub-block of dimension 256. Finally, a two-layer Transformer encoder is employed to effectively extract temporal features from the haptic signals.

\subsection{Performance Metrics}
In this study, our evaluation criteria for user identification and task classification performance are \gls{acc} and \gls{prec}, which measure the model’s ability to correctly identify users or tasks. 

\begin{enumerate}
    \item \textbf{Accuracy:} 
    Accuracy is an overall indicator of a model's correctness, calculated as the ratio of accurately classified instances to the total number of instances.

    \item \textbf{Precision:} 
    For a particular user, precision indicates the proportion of instances predicted to belong to that user which actually belong to that user.
\end{enumerate}

\section{Performance Evaluation}
\label{sec:Performance Evaluation}

\subsection{Task classification}
\begin{figure}[!t]
  \centering
  \vspace{2mm}
  \includegraphics[width=0.75\linewidth]{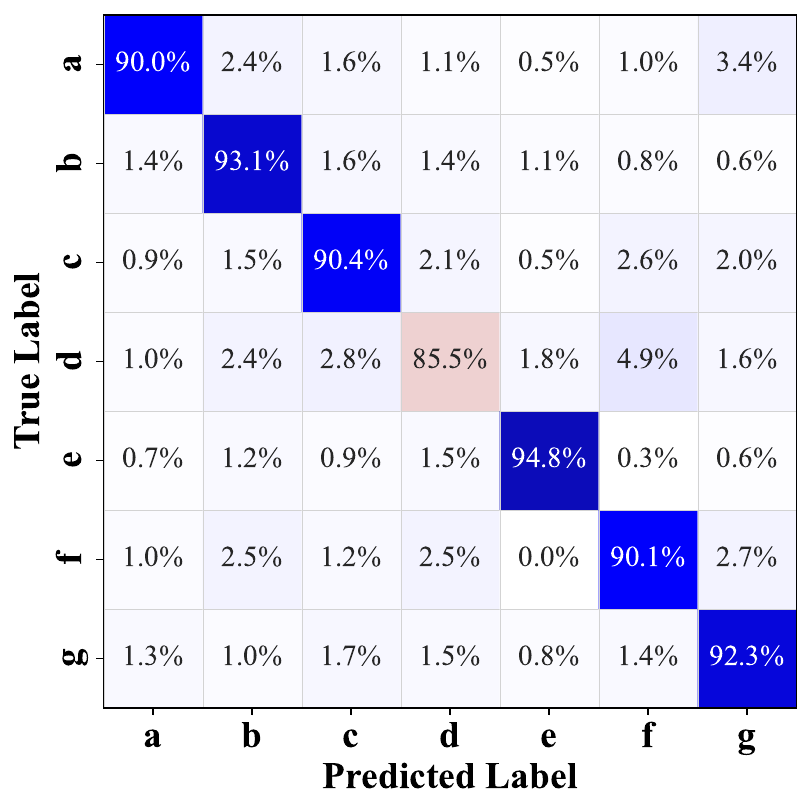}
  \caption{Confusion matrix of task classification (raw force).}
  \label{fig:task-force-raw}
\end{figure}

\begin{figure}[!t]
  \centering
  \vspace{2mm}
  \includegraphics[width=0.75\linewidth]{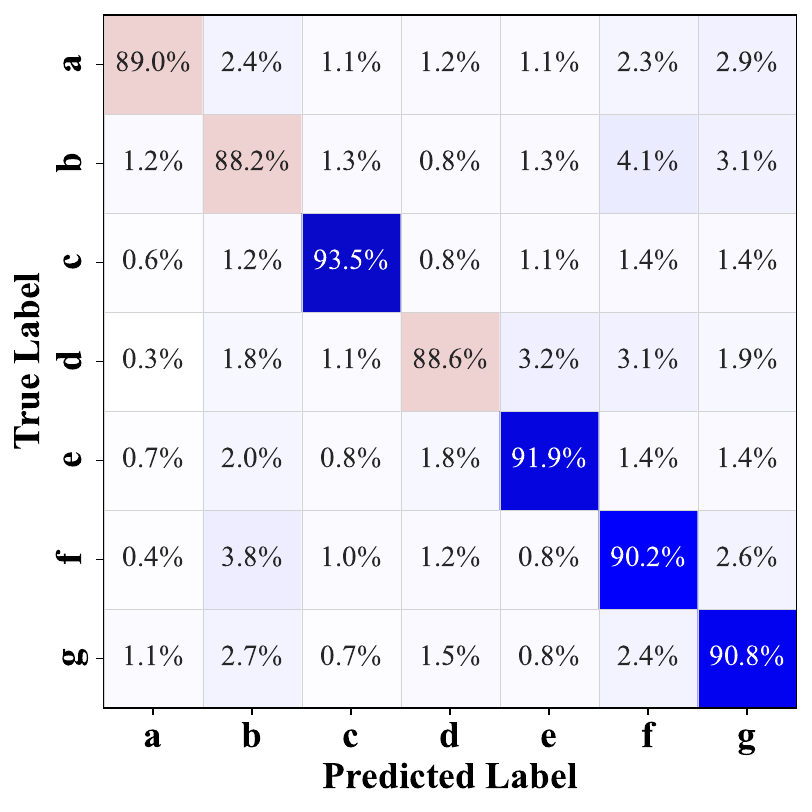}
  \caption{Confusion matrix of task classification (filtered force).}
  \label{fig:task-force-filtered}
\end{figure}

In this section, we evaluate the performance of task classification using haptic signals while the human operator performs different tasks. As demonstrated in our experimental results, both the raw force data and the filtered force data achieve high accuracy, exceeding 90\%.

This finding demonstrates that when a user manipulates a robotic arm, the force signals not only reflect the user’s intent but also capture the specific characteristics of the task at hand. Consequently, these haptic signals can serve as a reliable input for task classification in human-robot collaboration settings. However, given the potential privacy and security implications arising from the leakage of such data, it is crucial to ensure robust protection of these haptic signals.

\subsection{User Identification}

In this section, we evaluate user identification performance using haptic signals obtained as the human operator performs various tasks. 

\begin{figure}[!t]
  \centering
  \vspace{2mm}
  \includegraphics[width=0.9\linewidth]{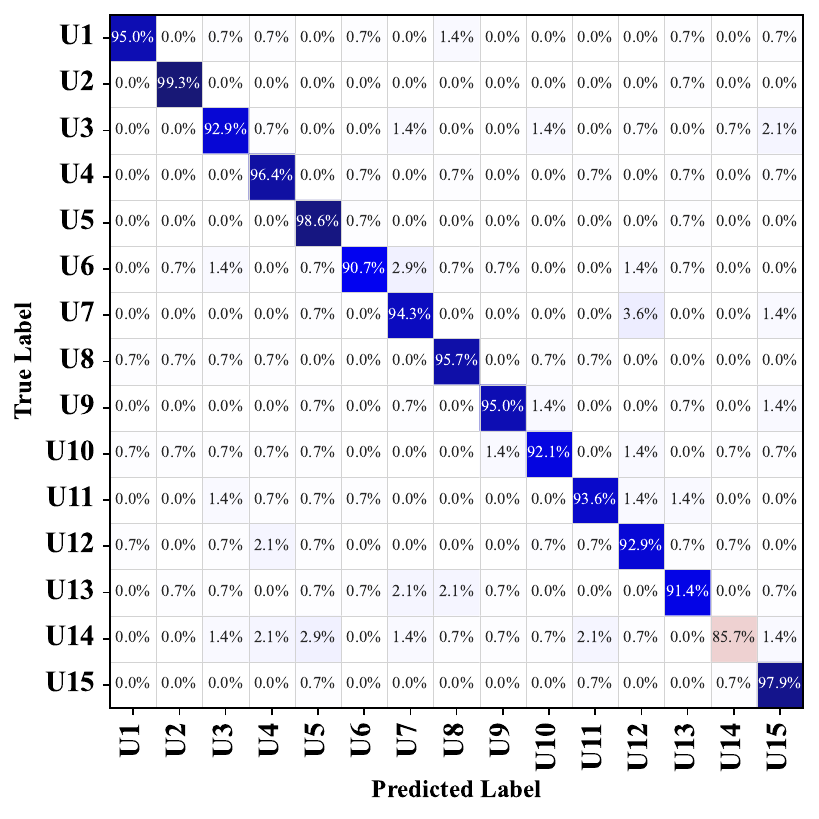}
  \caption{Confusion matrix of user identification (raw force).}
  \label{fig:user-train_raw}
\end{figure}

\begin{figure}[!t]
  \centering
  \vspace{2mm}
  \includegraphics[width=0.9\linewidth]{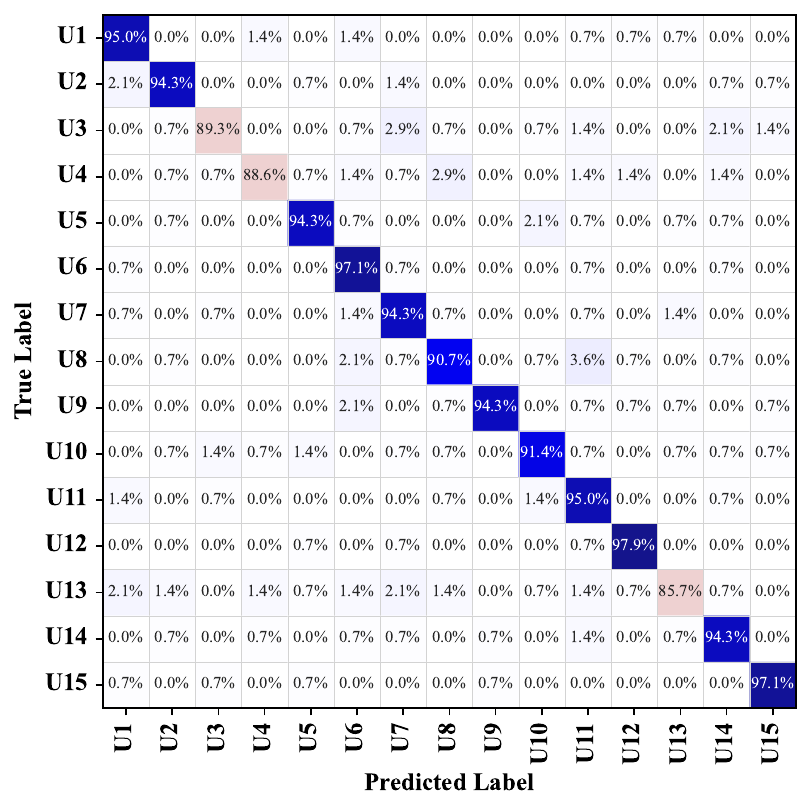}
  \caption{Confusion matrix of user identification (filtered force).}
  \label{fig:user-train_filtered}
\end{figure}

As shown in Fig.~\ref{fig:user-train_raw}, the raw force data yields high user identification rates (over 90\%) for all users. Meanwhile, the filtered force data (Fig.~\ref{fig:user-train_filtered}) achieves more than 90\% for every user except U3.

These results demonstrate that when an operator remotely controls a robotic arm, the forces they apply exhibit unique, individual force signatures. Consequently, force-based signals offer a promising approach for user identification in human--machine interaction. In effect, the robotic system captures and reflects human behavior through these force measurements, embedding the operator’s distinctive behavioral patterns into the robot’s behavior.

However, the feasibility to identify users based on their force signatures also introduces security and privacy concerns. Because such force data could be used to track or profile individuals, it is essential to establish robust data protection measures, such as anonymizing force signals, employing secure storage and transmission protocols to safeguard the privacy of operators.

\begin{figure}[!t]
  \centering
  \vspace{2mm}
  \includegraphics[width=0.70\linewidth]{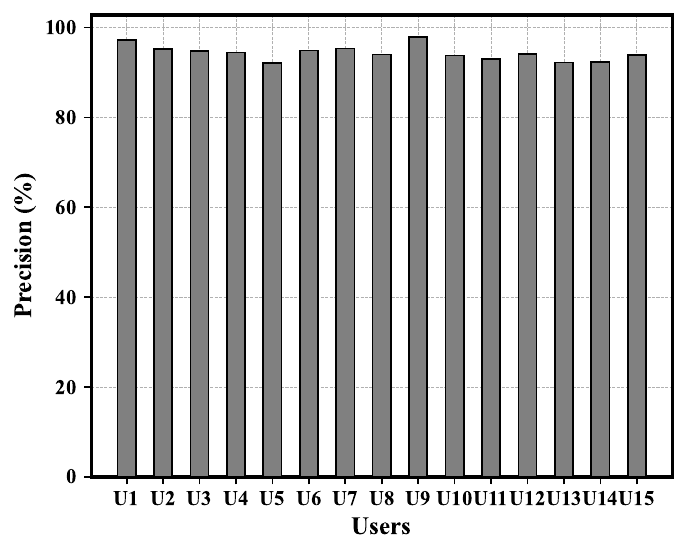}
  \caption{Performance user identification (raw force).}
  \label{fig:presicion_force}
\end{figure}

As shown in Fig.~\ref{fig:presicion_force}, we provide the average precision of each model across seven tasks. For user identification performance, using the raw force for recognition yields accuracies ranging from 90.27\% to 97.21\%, with an average of about 93.46\%. In comparison, using filtered force yields accuracies from 88.46\% to 98.58\%, also averaging around 92.89\%. 

These results demonstrate consistently high recognition accuracy for user identification, indicating that our method effectively leverages force-based haptic signals, which exhibit distinctive user uniqueness.

\begin{figure}[!t]
  \centering
  \vspace{2mm}
  \includegraphics[width=0.70\linewidth]{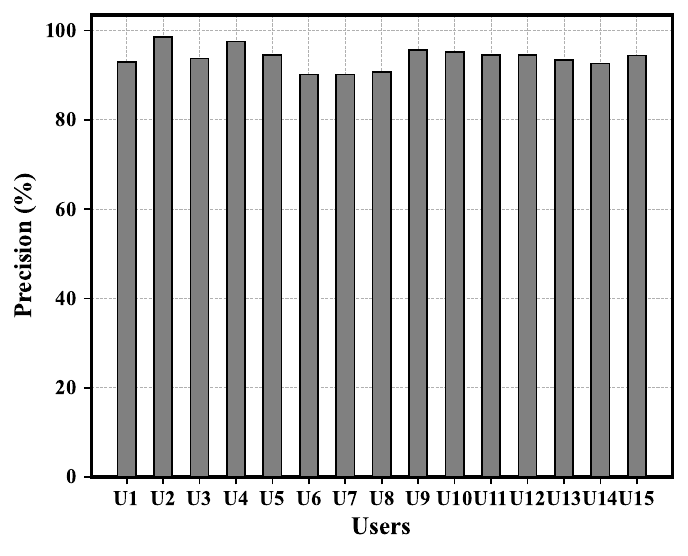}
  \caption{Performance of user identification (filtered force).}
  \label{fig:presicion_filtered_force}
\end{figure}

\subsection{Impact of Training Data Size}

\begin{figure}[!t]
  \centering
  \vspace{2mm}
  \includegraphics[width=0.75\linewidth]{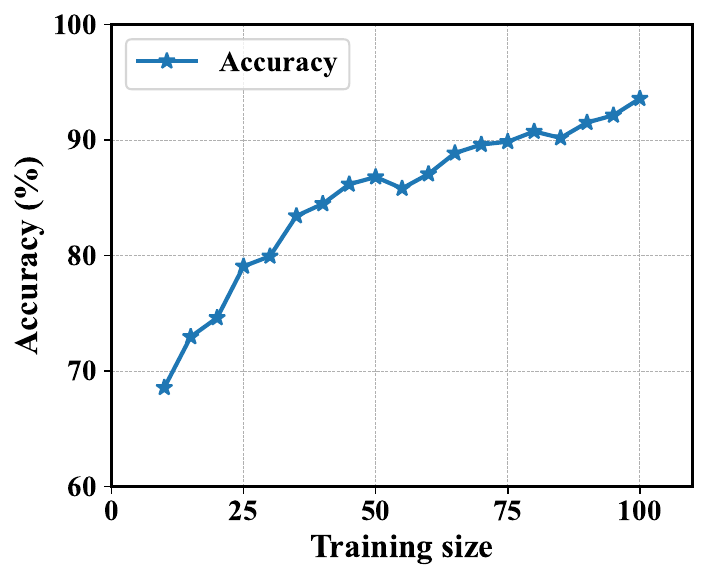}
  \caption{Performance of task classification under different training size (raw force).
  }
  \label{fig:train_size_force}
\end{figure}

\begin{figure}[!t]
  \centering
  \vspace{2mm}
  \includegraphics[width=0.75\linewidth]{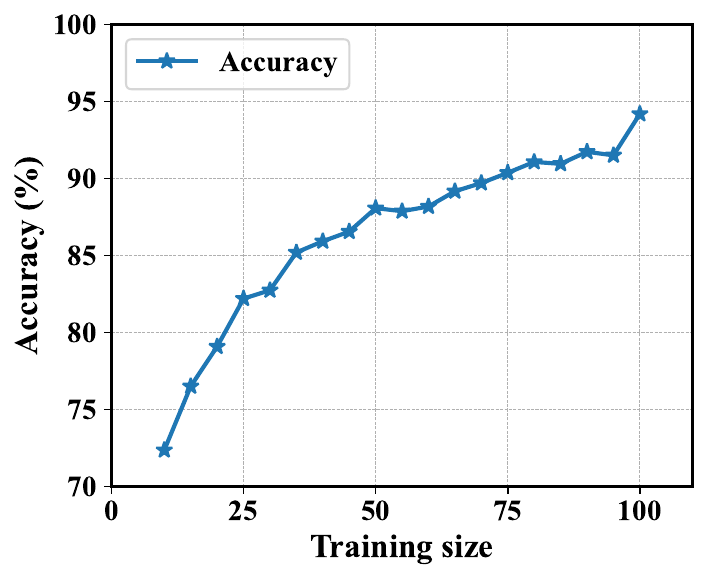}
  \caption{Performance of task classification under different training size (filtered force).}
  \label{fig:train_size_filtered_force}
\end{figure}

In order to investigate how varying the training data size affects task classification performance, we incrementally increase the number of training instances from 5 to 100 in steps of 5. As shown in Fig.~\ref{fig:train_size_force}, we present the average accuracy across different users using raw force data. Our results demonstrate that accurate task classification is achievable using haptic signals even with a limited amount of training data. Furthermore, the performance gradually improves as the training set size increases, ultimately exceeding 94\%.

We also explore task classification performance using filtered force data. As illustrated in Fig.~\ref{fig:train_size_filtered_force}, with just 5 training samples, the classification accuracy is around 72\%. Furthermore, this accuracy continues to improve as the training set size grows and reaches 95\%.

\section{Conclusion}
\label{sec:Conclusion}
This paper explores the performance of haptic signals for both user identification and task classification in a telerobotic human--robot interaction scenario. Our results indicate that force feedback inherently carries personal information about the operator, particularly in remote-control settings. Moreover, by employing a two-layer Transformer model architecture, we achieve over 90\% accuracy in both user identification and task classification. Our findings indicate that force feedback data, collected while a human operator controls a robotic arm, is inherently identifiable, raising significant security and privacy concerns. Consequently, our future work will focus on developing privacy-preserving methods to anonymize user data, investigating how to balance usability with robust personal information protection.

\bibliographystyle{IEEEtran}
\bibliography{bib}

\end{document}